\documentclass[a4paper, conference]{IEEEtran}
\usepackage{graphicx}
\usepackage{tabularx}
\usepackage{mathtools}

\usepackage{algorithm}
\usepackage{algorithmic}
\usepackage[numbers,sort&compress]{natbib}

\begin{document}

\title{Bionic Sea Urchin Robot \\ with Foldable Telescopic Actuator}



\author{\IEEEauthorblockN{Luis A. Mateos }
	\IEEEauthorblockA{MIT\\
	}
	
}

\maketitle
\thispagestyle{empty}
\pagestyle{empty}


\begin{abstract}


This paper presents a couple of interesting ideas: a telescopic actuator design and a bio-inspired sea urchin robot. The "spines" of the sea urchin robot consist of fourteen telescopic actuators equally distributed over it's spherical body. 

The telescopic actuation system integrates linked 3D  printed rack articulations that are locked in all axes if the pinion moves the links in a forward direction, creating a solid and rigid rack. Thus, the robot is able to propel and move by extending its spines.
On the other hand, if the pinion moves the rigid articulations backward, these are unlocked and can be folded in a minimal space, enabling the bionic sea urchin robot to hide its spines inside its constrained spherical-body and be able to roll.

Simulations and experiments are presented from both, the sea urchin robotic prototype and different scales of the telescopic actuation system.

\end{abstract}

\section{Introduction}
\label{sec:introduction}


Sea urchins have a round shaped body and with long spines that come off it. The spines of the sea urchin are used for multiple purposes, such as protection, to move about, and to trap food particles that are floating around in the water. Sea urchins are animals with little mobility as they move slowly with their moveable spines and typically range in size from $3$ to $10cm$ \cite{bookbarnes}.

\subsection{Hybrid spherical robots}

Spherical mobile robots embed a special morphology that has multiple advantages over common legged and wheeled robots: their outer shell protects them and their motion is smooth with good power efficiency. In addition, these robots are omni-directional,  they can move in any direction as any part of their outer shell can be considered as a foot, making them easy to recover after a collision and automatic adapting to soft or uneven terrains \cite{Armour2006}.

In the literature there are multiple modifications of spherical robots to make them more adaptable to different terrains and environments, such as spherical robots able to dive, swim, and even walk. Most of the spherical robots integrate a sealed shell and therefore they can jump in the water and swim from their rotational movement \cite{Suomela2005BallShapedRA}\cite{security}\cite{robotics1010003}. In addition, some spherical robots integrate a rugged sealed shell that increases their grip to the terrain and speed when swimming \cite{rotundus}. Also, there has been special designs of spherical amphibious robots. Li \cite{6618080} presented a re-configurable sphere with water-jets and four legs, enabling the robot to walk and dive. But lacking the ability to roll on terrain.


In the state of the art, there are no hybrid bionic robots similar to the proposed sea urchin robot. The main reason is the challenging mechanical actuation system, which requires a strong and fast extension/compression of the telescopic levels able to lift the robot when extending, and able to fit the compressed actuator inside the robot's spherical body.

\begin{figure}[t]
	\centering
	\includegraphics[width=0.49\textwidth]{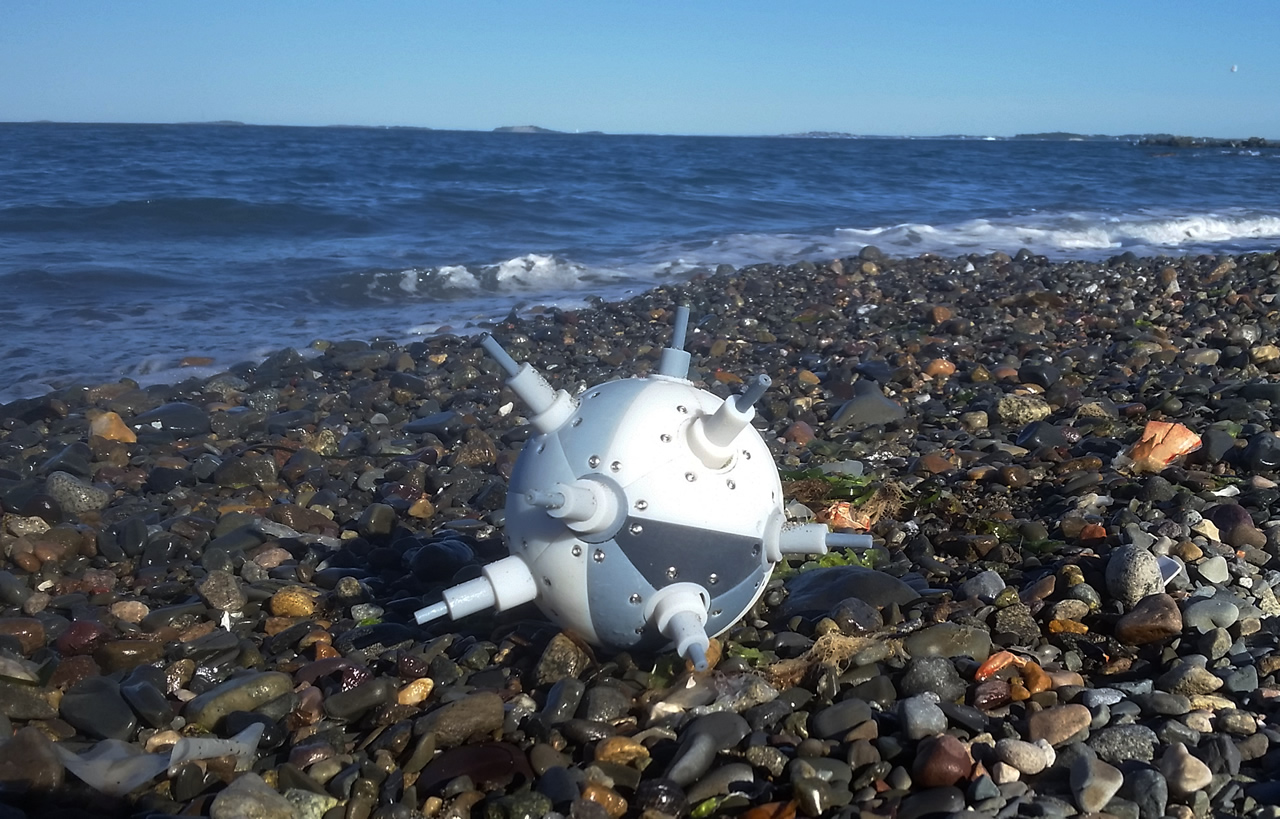}
	\caption{Bionic sea urchin robot. }
	\label{ffig1}
\end{figure}

\begin{figure*}[t]
	\centering
	\includegraphics[width=0.99\textwidth]{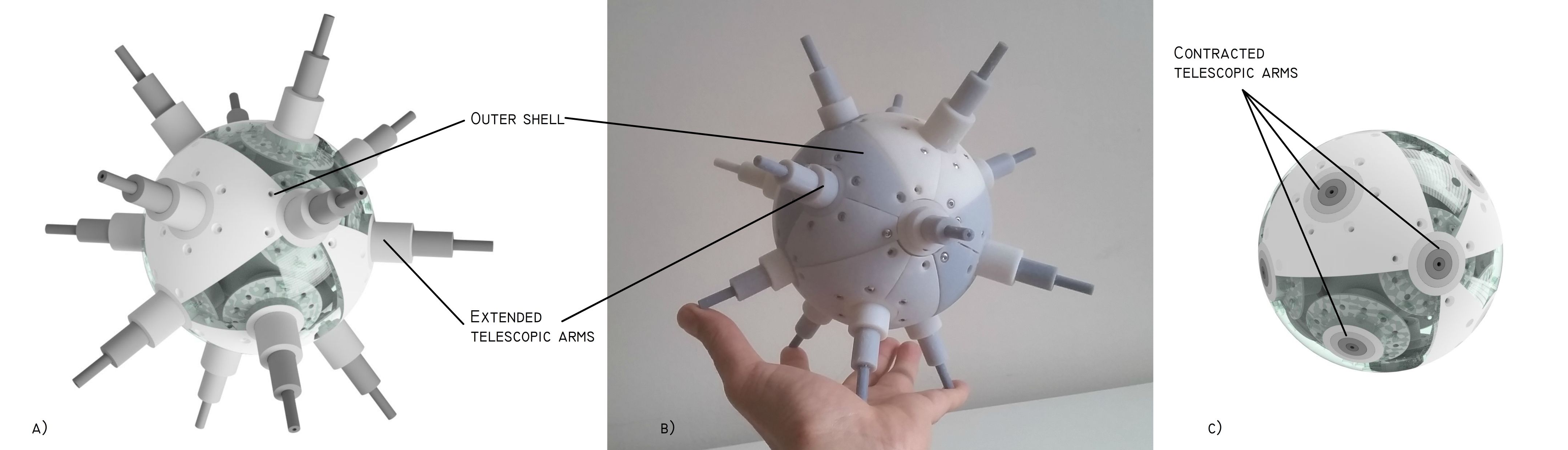}
	\caption{a) Sea urchin robot (3D model) with extended telescopic actuators. b) Bionic sea urchin (real-robotic prototype) with extended telescopic actuators. c) Sea urchin robot (3D model) with compressed telescopic actuators, mimicking a spherical mobile robot.}
	\label{figtelescopic1}
\end{figure*}

\subsection{Electric telescopic actuators}

Electric linear actuators generally offer a low ratio between the length $L$ when fully extended to the length $l$ when compressed, commonly a fully extended actuator is less than twice the size when fully compressed $L<2l$. The reason is that in its body rest a rigid piston which can not be folded neither bended \cite{sclater2007mechanisms}. 

However, there has been other mechanical methods in electric linear actuators that can extend considerably while reducing their size when compressed. 
One example is the rigid chain actuator \cite{chainlink}. This is a specialized mechanical linear actuator used for push-pull lift applications. The actuator is a chain and pinion device that forms an articulated telescoping member to transmit traction and thrust. The links of the actuating member are linked in a way that they deflect from a straight line to one side only. As the pinions spin, the links of the chain are rotated $90^\circ$ through the housing, which guides and locks the chain into a rigid linear form effective at resisting tension and compression (buckling). In this way, the actuating member can be folded and stored compactly, either in an overlapping or coiled arrangement. 
However, one drawback of this mechanism is that it cannot support forces acting perpendicular to the axis of extension, since these will bend it in the same way as it is stored.

An improved version of the rigid chain actuator is the zip chain actuation unit, which interlocks a couple of articulated chains in a zipper-like fashion to form a single, strong column that enables push/pull operation with a maximum speed of 1000$mm/sec$. 
However, this actuation system requires a couple of articulated chains  \cite{zipchainlink}. 



Another highly compressible linear actuator is the  Spiralift \cite{spiralift}, which is a system of interlocking horizontal and vertical metal bands that "unroll" to lift a load, this is the most compact electric linear actuator \cite{7487363}. The spiralift can be integrated as a robot's vertical spine (eg. on EL-E \cite{Jain2009}, PR2\cite{pr2} and RED1\cite{qwerty}) -- increasing the robot's effective workspace, and also for compact and easy transportation. 
Regardless of its highly compressibility, the spiralift is slow to extend (max. speed of $10mm/s$) if compared to direct drive actuators, such as rigid chain actuator. The reason of the low speed is the interlocking process that rotates a band to form the tube. Nevertheless, this mechanism is able to lock in all axes, and  withstand forces at any point of the extension.

\begin{table}[b]
	\centering
	\caption{Highly extendable linear actuators characteristics. }
	\begin{tabular}{|l|l|l|l|l|}
		\hline
		 & \textbf{articulated }  & \textbf{speed}     & \textbf{locked}  \\ 
		 & \textbf{ chain}  & \textbf{mm/s}    & \textbf{axes }  \\ \hline
		\textbf{Rigid chain}             & 1     & fast ($1000mm/s$)   & 2         \\ \hline
		\textbf{Rigid zip }              & 2     & fast  ($1000mm/s$)    & 3         \\ \hline
		\textbf{Spiralift}               & 1     & slow  ($10mm/s$)    & 3         \\ \hline
		\textbf{Articulated rack}       & 1     & medium  ($100mm/s$)    & 3         \\ \hline
	\end{tabular}
\label{tabtable}
\end{table}

In this paper, a bionic sea urchin is presented with the capabilities to extend and compress its spines for swimming, moving over rocks and overcome obstacles, see Fig. \ref{ffig1}. One of the challenges to develop this bionic robot is to integrate as many telescopic actuators as possible with the capability to compress them inside the spherical body of the robot so it can become a sphere and roll. Another challenge is to have the capability to extend the linear actuators considerably, about half of its spherical body diameter to lift up the robot. 

The novel telescopic linear actuator is based on the rack and pinion principles, in which a geared motor with the pinion rotates and moves a rack forward and backward, pushing or pulling the actuator's levels. However, instead of integrating one piece of rigid rack, the actuation system integrates articulated racks that are link together. If the racks are moving in the forwards direction, the links are locked in all axes, creating a rigid rack able to push/pull, while overcoming forces perpendicular to its axis of extension at any point. If the rigid articulated rack is moving backwards, the rigid rack is unlocked becoming an articulated chain of rack-links that can be folded and stored in a minimal space inside the robot's body. Hence, the proposed foldable actuation system requires only one articulated chain which locks the links in all axes with a speed of $100mm/s$, see Table \ref{tabtable}.

This document is structured as follows: 
Section \ref{sec:design} presents the design of the sea urchin robot and its relevant features.
Section \ref{sec:model} describes the design of the telescopic actuators and the articulated rack and pinion gearbox. 
Section \ref{sec:simulation} shows simulation results from the robot's locomotion.  
Section \ref{sec:prototype} presents the robotic prototype. 
Conclusions are proposed in Section \ref{sec:conclusion}. 



\section{Sea Urchin Robot Design}
\label{sec:design}


The presented robotic sea urchin can be remote control or semi-autonomous. 
The robotic system includes: xbee modules, a couple of microcontrollers ATmega328, seven DRV8835 dual motor driver and one BNO055 absolute orientation sensor. In this configuration, all 14 telescopic actuators can be controlled, while registering the inclination and orientation of the robot, see Fig. \ref{figtelescopic1}. In the semi-autonomous mode, the robot can be set for basic tasks: such as move straight  or in a curvilinear fashion and perform  preprogrammed movements.






The outer shell of the robot protects the electronics, motors and holds the telescopic actuators. It consists of 24 3D printed parts that forms a sphere with diameter $D=130mm$ and holds 14 telescopic actuators. The material used for printing the shell is a mix of 3D print-materials (RGD450 as a rigid and TangoBlack as a flexible material) with a tensile strength of $30-35MPa$ and elongation at break 50-65\%. In this setup, the robot can withstand hits when rolling or falling from uneven surfaces.

\begin{figure}[t]
	\centering
	\includegraphics[width=0.45\textwidth]{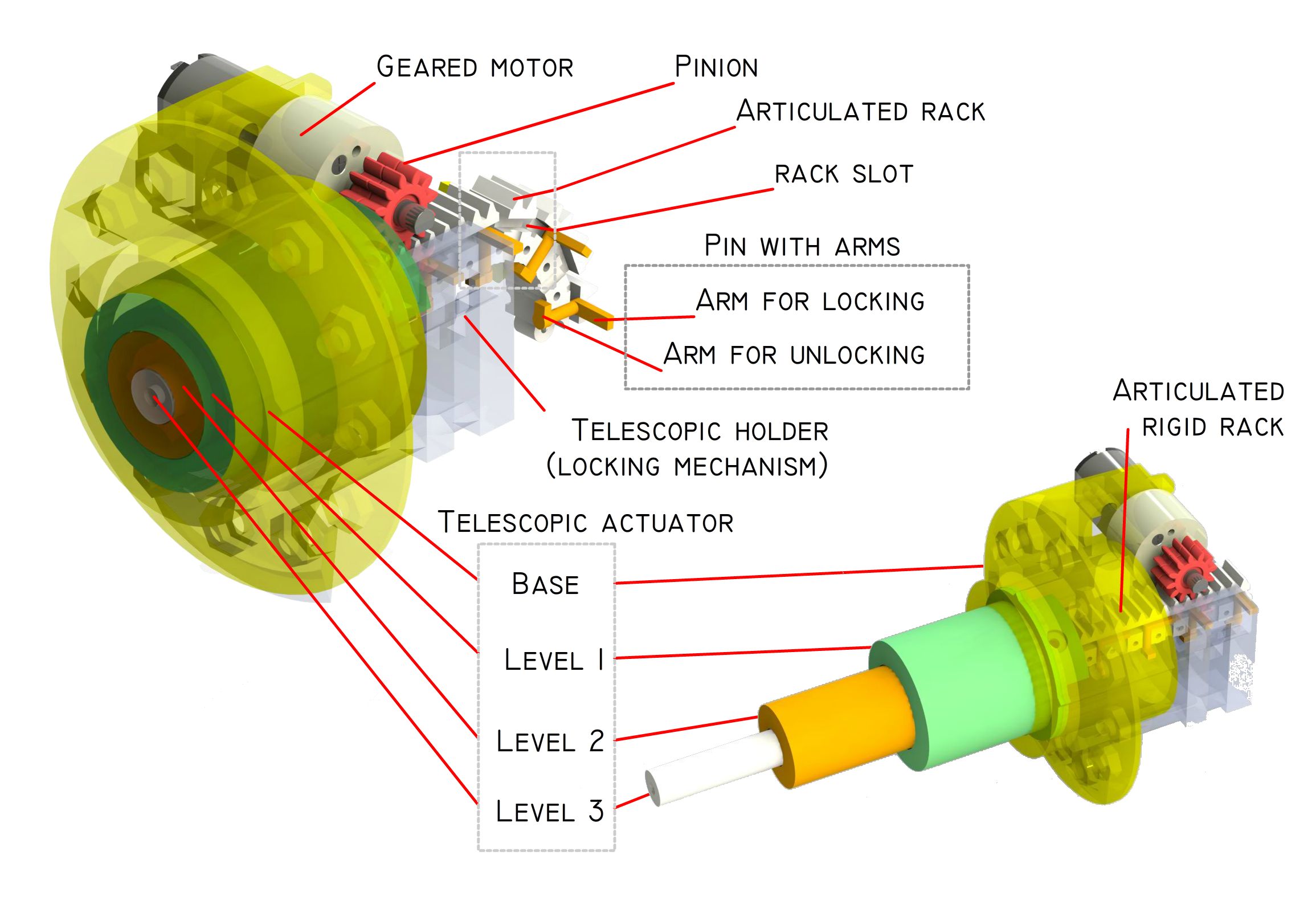}
	
	\includegraphics[width=0.45\textwidth]{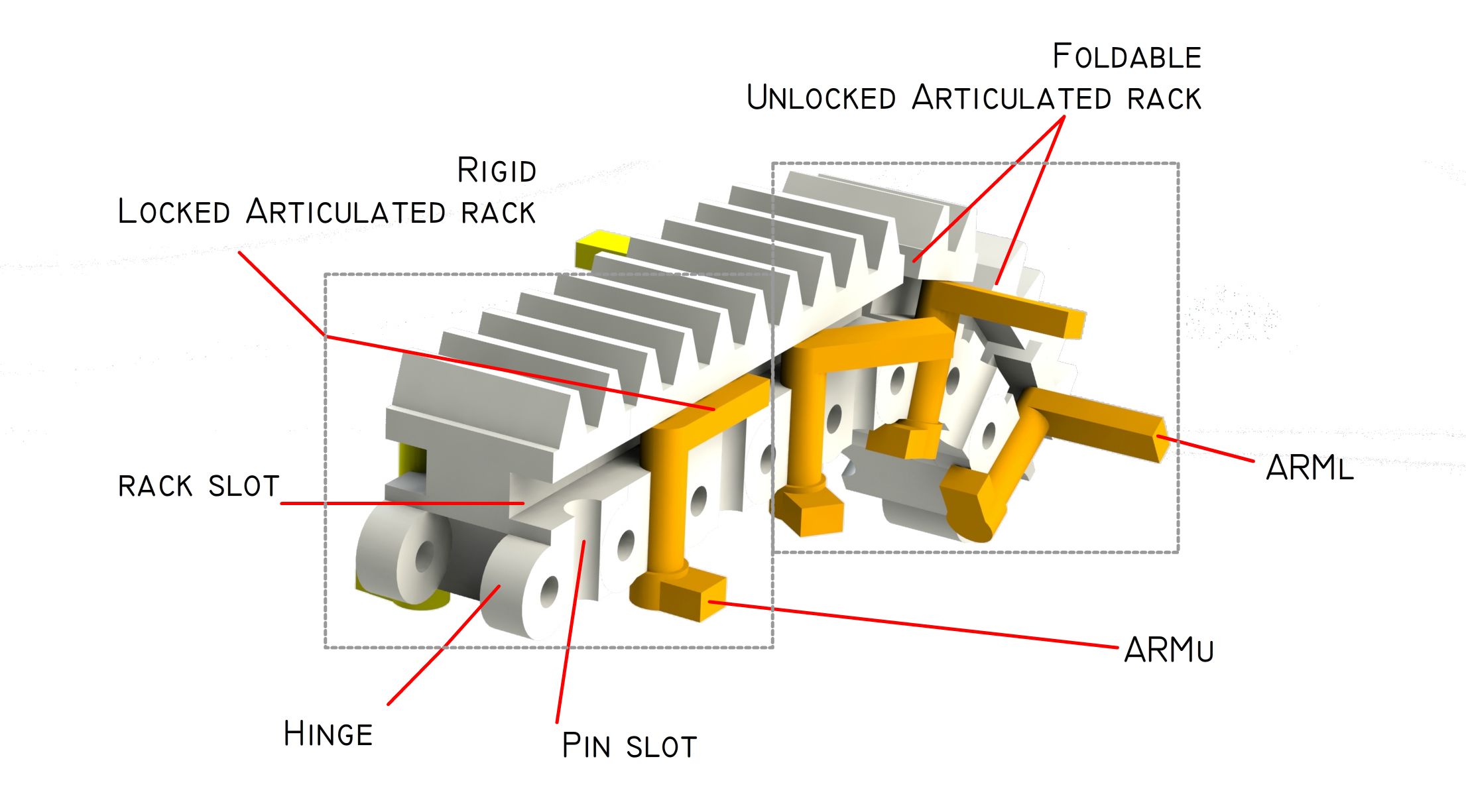}
	\caption{ Telescopic actuator contracted and extended (top). Articulated rack components. If the arm $ARM_L$ is set inside the rack slot, then the articulated rack becomes rigid. If the arm $ARM_L$ is rotated and set out of the rack slot, then the rack becomes articulated and can be folded (bottom).}
	\label{figteles3}
\end{figure}

\section{Foldable Telescopic Actuator Design}
\label{sec:model}

%

The robotic sea urchin requires a novel telescopic actuation able to fit in a minimal space when compressed and with a fast and strong extension ratio to lift and propel itself. 

The novelty of the presented actuation system with articulated racks is that when extending it, the articulations are reconstructed as a solid rigid rack, locking the articulations in all axes. Whereas, when compressing the articulations, these are unlocked and can be folded. This development resembles latching systems for reconfigurable robots, such as in-pipe robots \cite{6766548} and autonomous robotic boats\cite{8793525}.

\subsection{Telescopic levels}

The telescopic actuator consists of four levels that can be extended up to $l=64mm$ from the top of its base, see Fig. \ref{figteles3}. The length of the actuator when extended is $L=89mm$ and when contracted is equal to the $base_{height}=25mm$ with an extension ratio of 1:3.56. 

The base level of the telescopic actuator is 3D printed with rigid material in order to support all the  telescopic levels and connections with the outer shell parts. 
The following three levels in the telescopic actuator are combinations of 3D printer materials: rigid $75\%$ and flexible $25\%$ in order to withstand hits, similar to the outer shell parts. 


\subsection{Articulated rack-gear}

The telescopic actuator integrates a 3D printed rack \& metallic pinion gearbox to extend and contract the levels. A geared motor attached to the telescopic's base level is the actuation unit "pinion", which drives the articulated rack attached to the top level of the telescopic arrangement. 

Each articulated rack consists of a couple of gear-teeth, a rack-slot on the sides, a pin with a couple of arms to lock/unlock the articulations and a hinge on both ends for linking the rack articulations, see Fig. \ref{figteles3}.

\begin{figure}[b]
	\centering
    \includegraphics[width=0.4\textwidth]{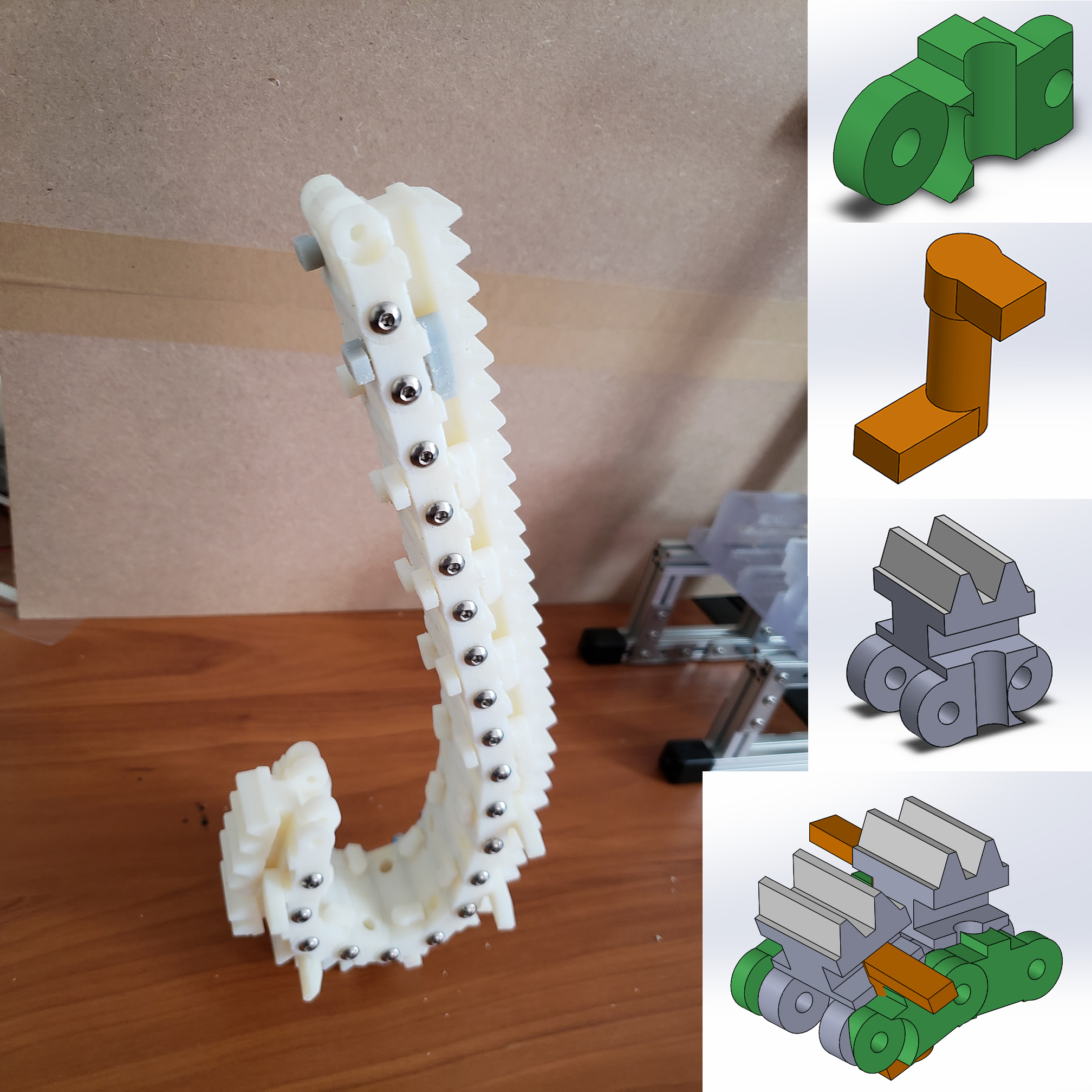}
	\caption{3D printed articulated rack prototype with locked (rigid) and unlocked (foldable) segments (left). Articulated rack components: Pin cover, cylindrical pin with arms, gear-teeth, 3D model of a couple articulations (right).}
	\label{figteles35}
\end{figure}

\begin{figure*}[t]
	\centering
	\includegraphics[width=0.99\textwidth]{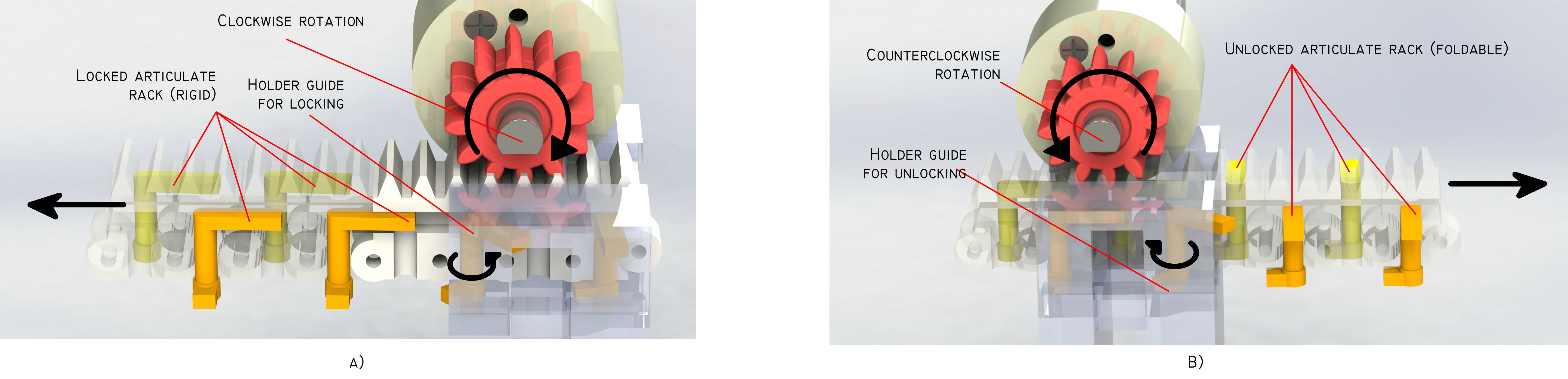}
	\includegraphics[width=0.32\textwidth]{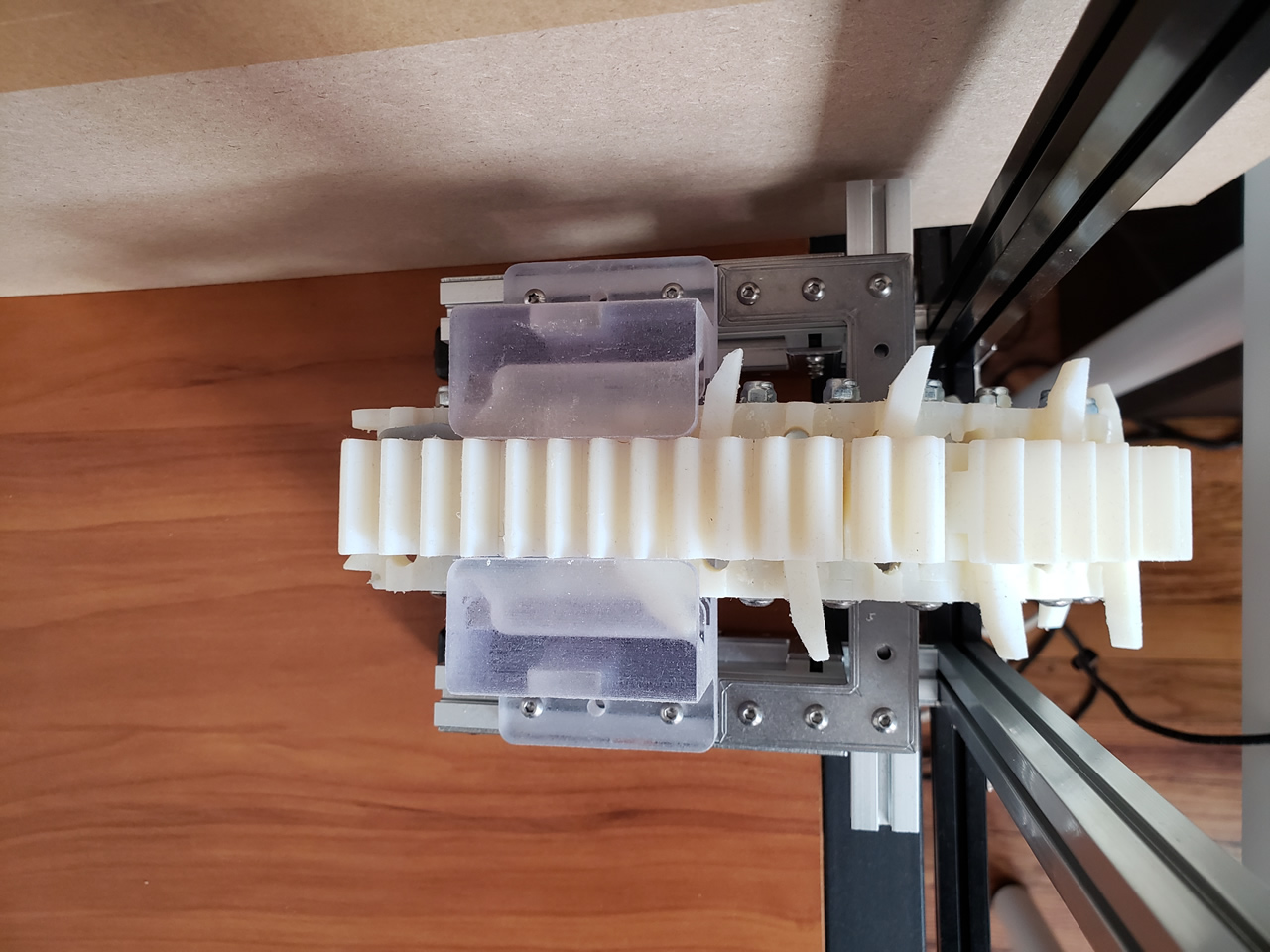}
	\includegraphics[width=0.32\textwidth]{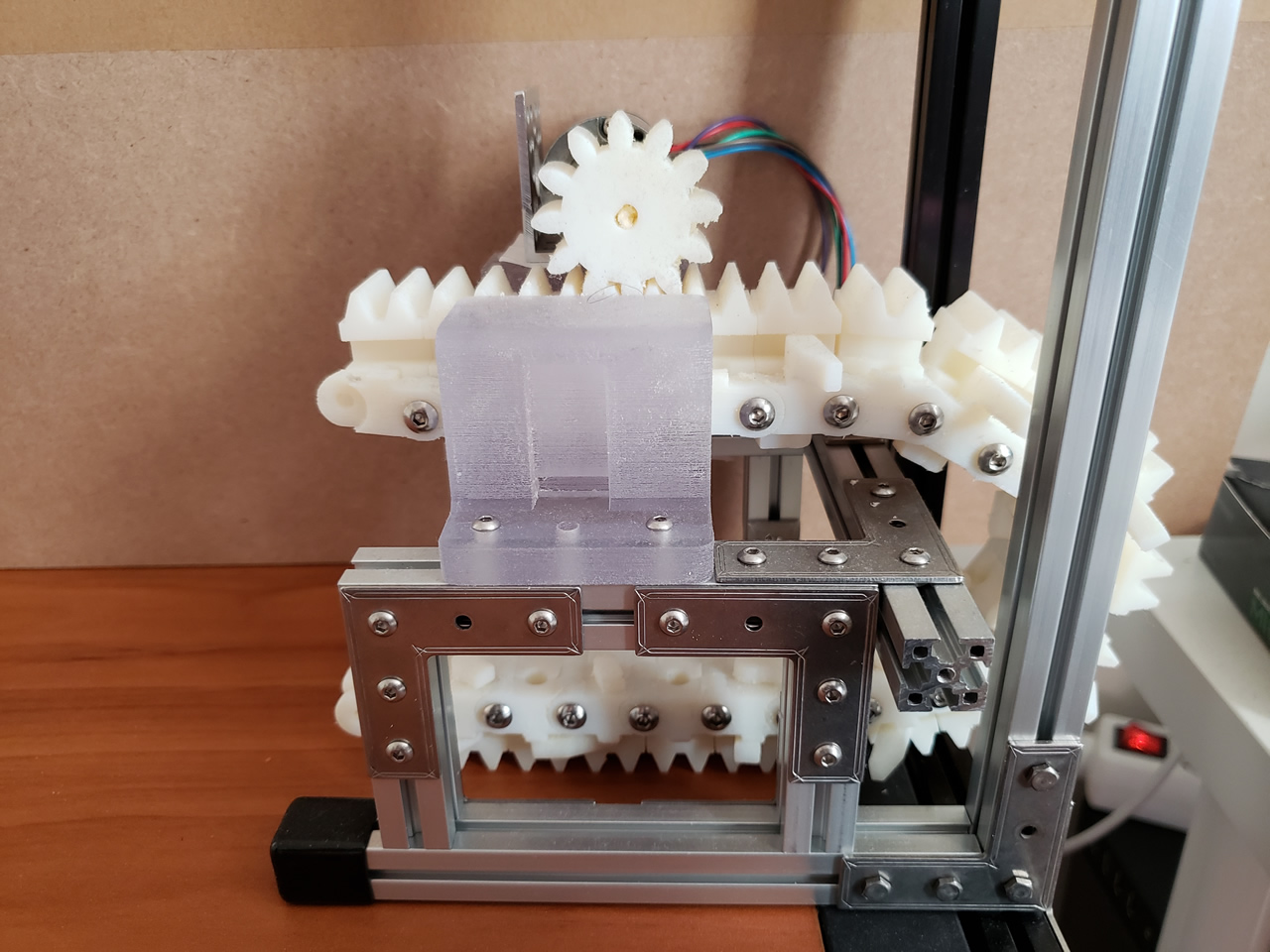}
	\includegraphics[width=0.32\textwidth]{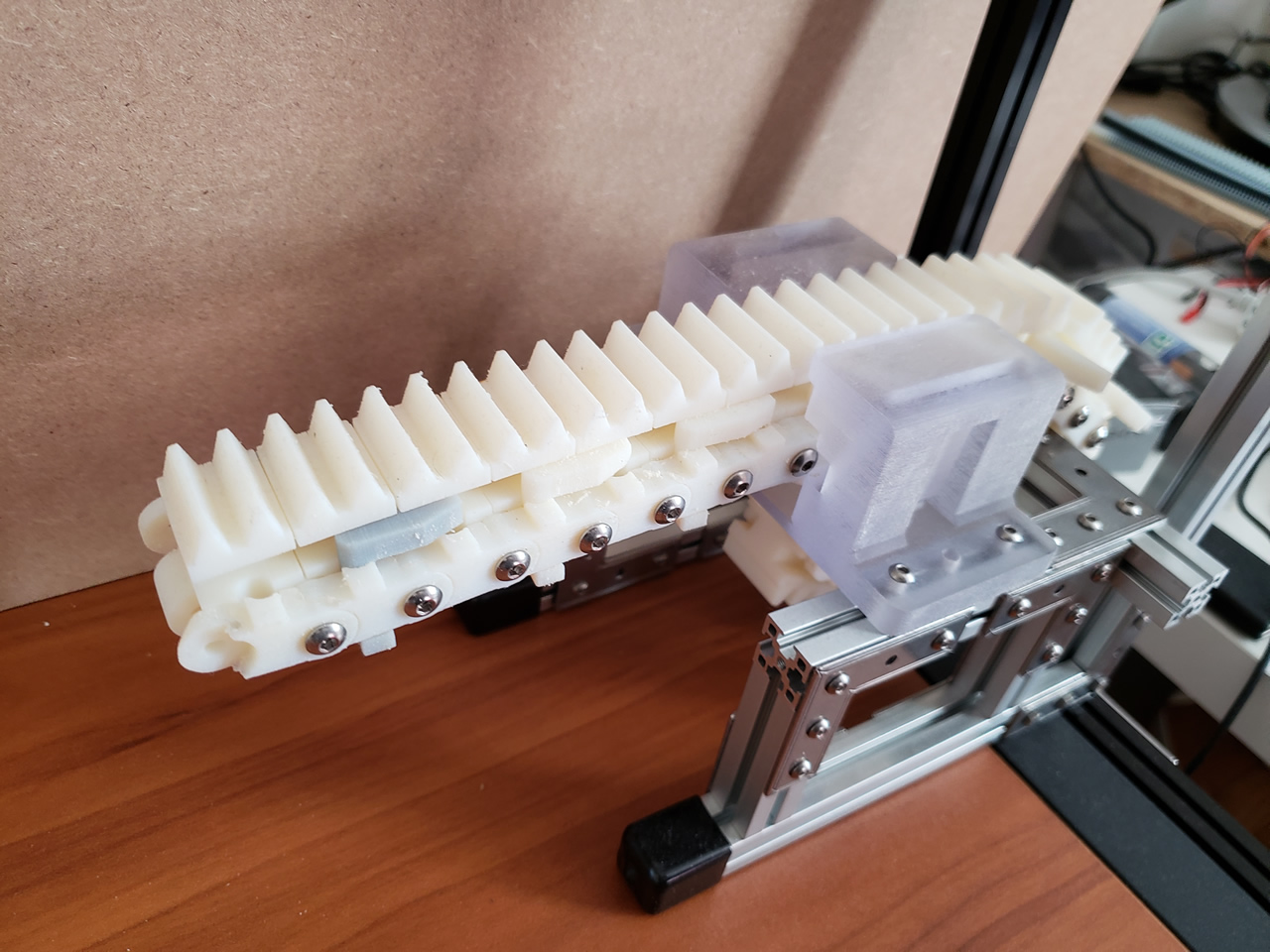}
	\caption{a) The articulated rack is changed from foldable to rigid when the geared motor (pinion) rotates clockwise, moving the articulated rack out from the telescopic actuator. The inner guides from the telescopic holder guide the arm $ARM_L$ to fit in the rack's slot. b) The articulated rack is changed from rigid to foldable when the geared motor (pinion) rotates counterclockwise, moving the articulated rack inside the telescopic actuator. The inner guides from the telescopic holder guide the arm $ARM_U$ to rotate and consequently moves the arm $ARM_L$ out of the rack's slot (top). 3D printed prototype with rigid and foldable parts before and after passing through the guides from the telescopic holder (bottom).}
	\label{figteles}
\end{figure*}

\subsubsection{Gear-teeth}

The articulations can be set with $n$ number of gear-teeth. However, when the articulations are folded, the more gear-teeth per articulation will lead to a lower degree of compression. 
The optimal solution for folding the articulated rack to a minimum is with a single gear-tooth per articulation. However, in the presented prototype, the articulations are set to a couple of gear-teeth. Since, it is easier to construct, arrange and have a good compression ratio.

\subsubsection{Rack-slot and pin with arms}

Each articulated rack has a rectangular and cylindrical slots on each side. 
A pin is inserted in the cylindrical slot, this pin consists of a cylindrical segment with a couple of arms, each  attached on each pin's edge. The arms are oriented $90^\circ$ from each other, one is always parallel to the rack extension axis and the other is perpendicular to it, pointing out from the rack gear. 
The arm $ARM_L$ is used for locking the articulated rack, while for unlocking the arm $ARM_U$ is used, see Fig. \ref{figteles35}. 



\subsubsection{Hinge on both ends}
The articulated racks integrate a hinge connector on both ends to interlock the articulations and create a longer rack. The hinge connectors are also use for attaching a cover to the cylindrical pin with arms, so it is always trapped when rotating.

\subsubsection{Telescopic holder}
Besides the telescopic base level which integrates the gear-motor, an extra holder is required to guide the articulated rack and transform it from rigid to foldable and vice-versa. The holder integrates a couple of guides, one for each arm of the pin ($ARM_L$ and $ARM_U$) for locking or unlocking the rack articulations.

\subsubsection*{Articulated rack locking}  If the pinion attached to the gear-motor rotates clockwise, it will move the articulated racks out from the actuator, and the telescopic holder will guide the pin's arm $ARM_L$ to fit inside the rack's slot, changing the articulation properties from foldable to rigid, while extending the telescopic actuator, see Fig. \ref{figteles}$a$. 

\subsubsection*{Articulated rack unlocking} If the pinion rotates counterclockwise, it will move the articulated racks inside the actuator, so the guides from the telescopic holder rotates the $ARM_U$ and consequently rotating the $ARM_L$ releasing it from the rack's slot,  changing the articulation properties from rigid  to foldable, see Fig. \ref{figteles}$b$.


\section{Simulation}
\label{sec:simulation}

\begin{figure*}[t]
	\centering
	\includegraphics[width=0.32\textwidth]{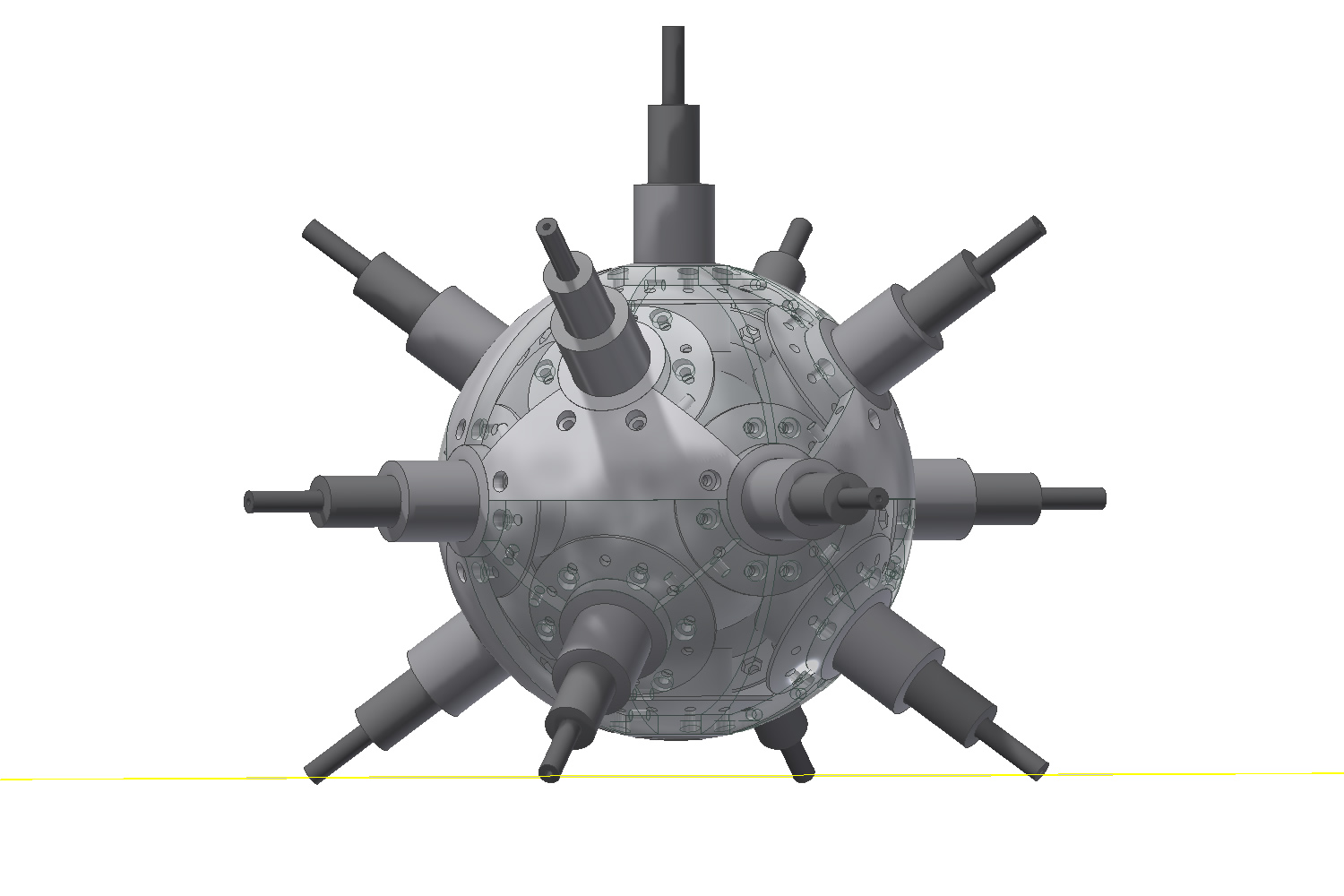}
	\includegraphics[width=0.32\textwidth]{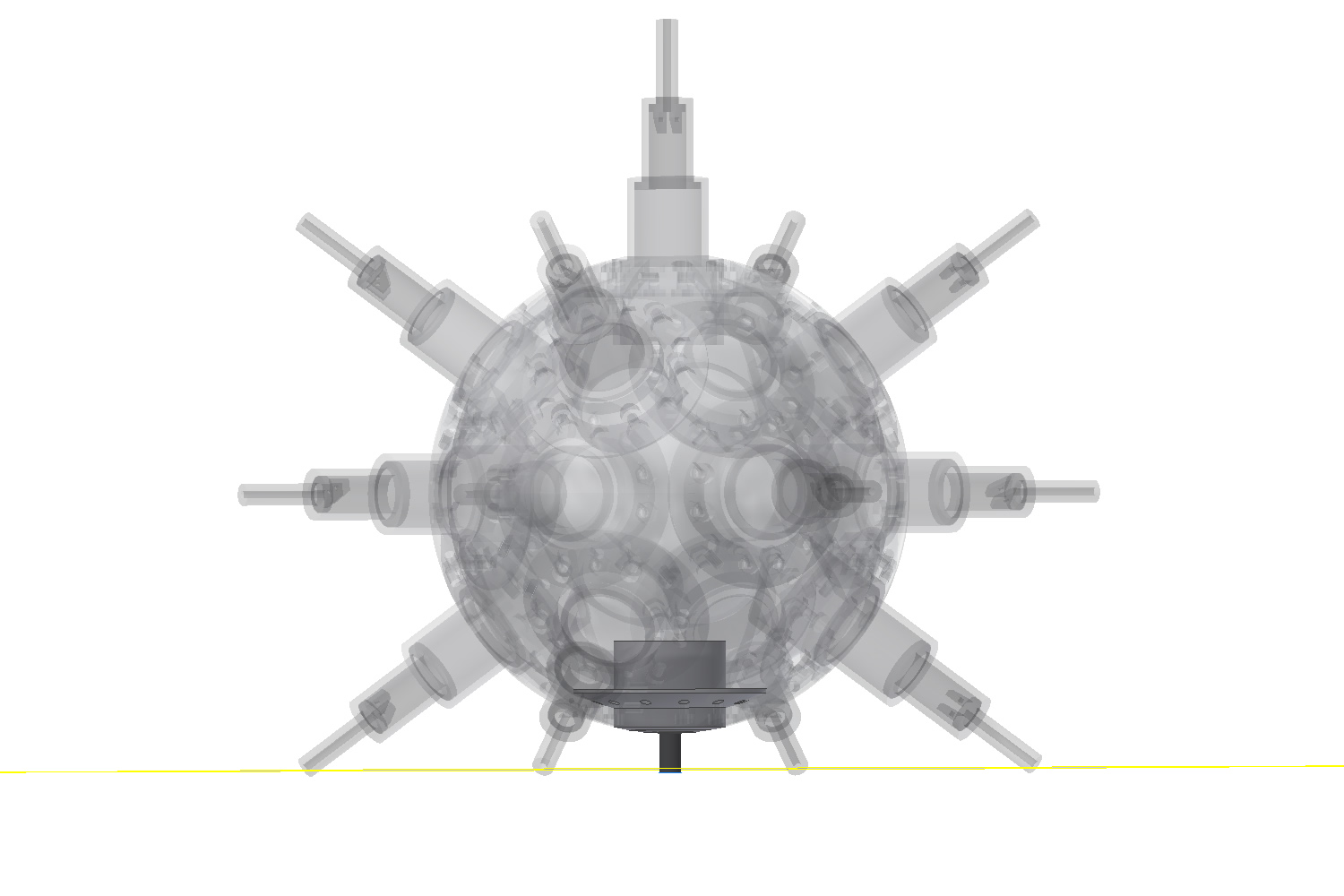}
	\includegraphics[width=0.32\textwidth]{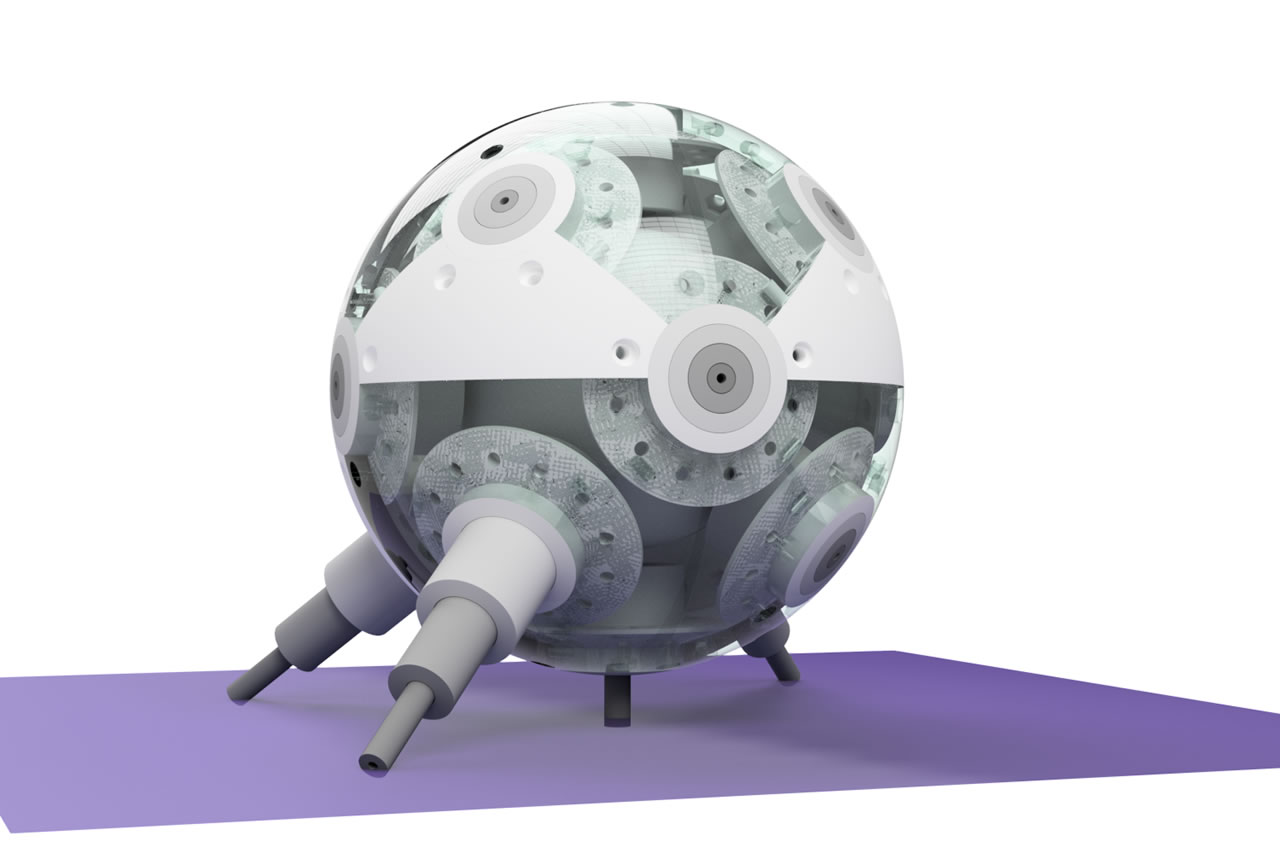}
	\caption{Robot in resting position with four spines (left), the fifth spine propels the robot (center) and the compressing spine sets the direction (right). }
	\label{figmodelx}
\end{figure*}

The robotic sea urchin integrates 14 linear actuators, eight of them  located in the center of each octave of the sphere and the other six are located on the joining points of these octaves, see Fig. \ref{figtelescopic1}. Hence, the bionic sea urchin robot is able stand with 3, 4 or 5 spines on a flat surface and can be propelled in multiple ways. In this paper, a couple of locomotion methods are analyzed: 

The first locomotive configuration starts when 
the robot is set with four of the spines extended equally and in contact with the ground (on flat surfaces). On rough surfaces (i.e., over rocks) the robot extends the four spines until all are in contact and the inclination level of the robot is secured. In this way, the robot can be set stable to the ground level or above it. The fifth spine which is facing the ground plane, is used for lifting the robot, see Fig. \ref{figmodelx}.

In this configuration, the robot can move from its starting position by extending the fifth spine facing the ground and by compressing one of the other four spines. This coupled movement creates a shifting in the center of mass that makes the robot to roll. 
The spine facing the ground helps propelling the spherical robot, while the spine that is compressing sets the direction for the robot to move, see Fig. \ref{figmodelx}.

\begin{figure}[h]
	\centering

	\includegraphics[width=0.43\textwidth]{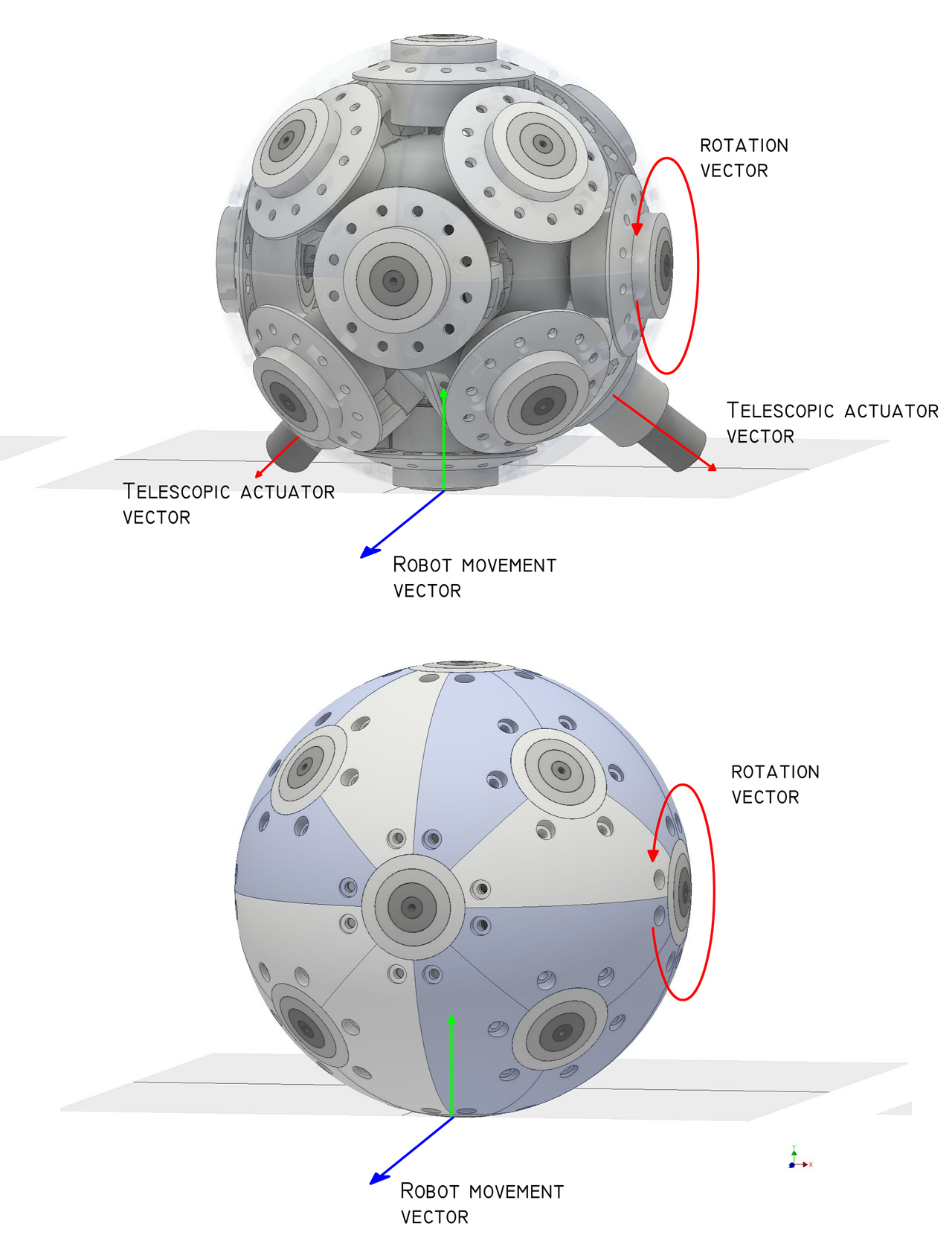}
	\caption{Particle robot in resting position. }
	\label{figmodel}
\end{figure}

The second locomotive configuration starts when all the spines compressed and with one spine facing the ground. In this case, the robot can propel by extending any two neighbor spines. 
In a flat surface, the robot can move towards a direction when a couple of neighbor spines (opposite to the direction to move) are extended at the same time, see Fig. \ref{figmodel}.

On the other hand, if the robot is rolling and wants to stop, it can extend the spines opposite from the rolling movement and stop immediately. 

The robot is not able to float, its buoyancy is low and once on water it sinks. Nevertheless, under the water the robot is also able to propel similar as on the surface. Plus, the robot is able to "jump" under the water when extending the spine facing the ground, see Fig. \ref{figprotosnow}.



\section{Experimental Results}
\label{sec:prototype}


The sea urchin robot has been tested in different challenging environments for common spherical mobile robots, such as snow, water, sand and rocks.

Common spherical mobile robots require a solid flat surface to propel itself and start rolling. The contact with the 
flat surfaces must have enough grip, so the robot's shell makes proper contact to the surface and the robot can roll efficiently. Otherwise, the spherical robots will be slipping on the surface without moving. This is the case of spherical mobile robots trying to move in challenging environments such as snow, ice, sand and rocks. 

On the other hand, the  bionic sea urchin robot is able to move and adapt on semi-rigid environments that are impossible for common spherical robots. Over icy snow the robot is able to move as well as over rocks. It can move on surfaces as long as the spines have a contact point to propel.

\subsubsection{Sand and rocks}
Spherical robots with polished shell are able to move on solid terrains but on dry sand they get stock, since there is no grip between the polished shell and the fine grains of sand. If the spherical robot integrates a rugged shell then it may move also on dry sand, overcoming small sand slopes \cite{rotundus}. Similarly, the bionic sea urchin robot is able to move on solid sand and small rocks by extending its spines, see Fig. \ref{figprotosnow}.

\subsubsection{Water environment}
\begin{figure*}[t]
	\centering
	\includegraphics[width=0.38\textwidth]{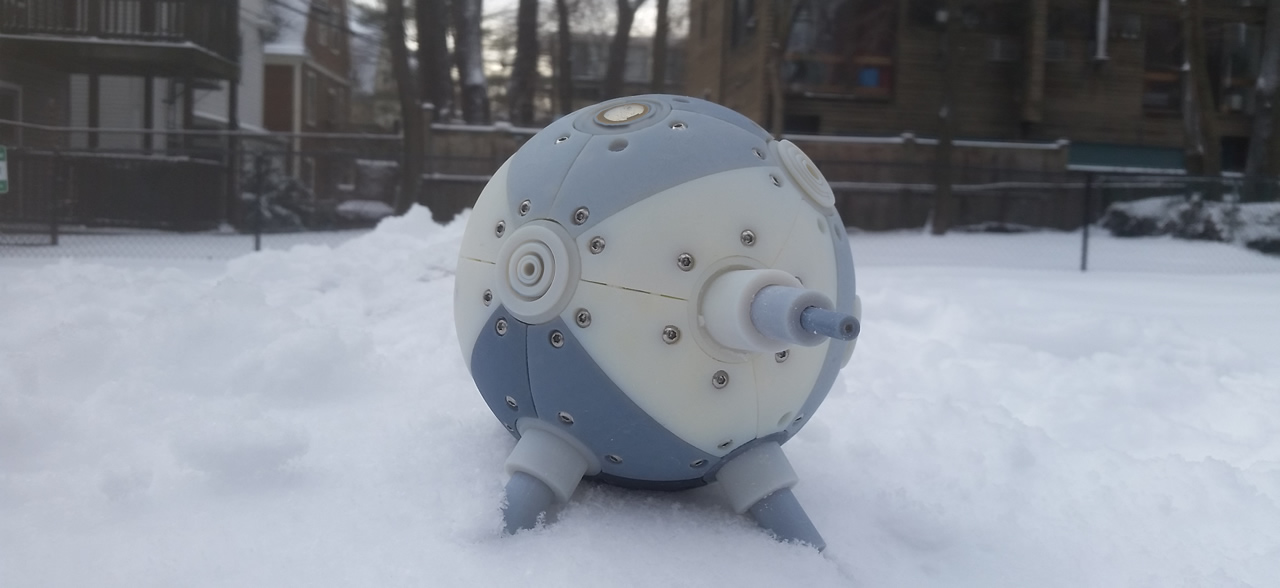}
	\includegraphics[width=0.38\textwidth]{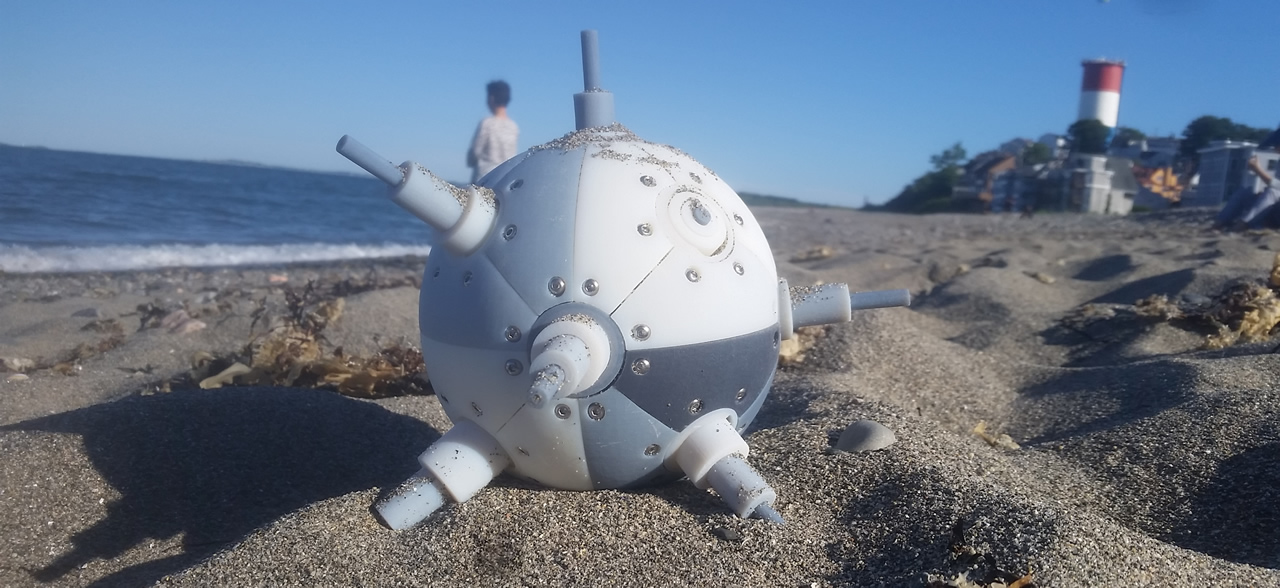}
	\includegraphics[width=0.38\textwidth]{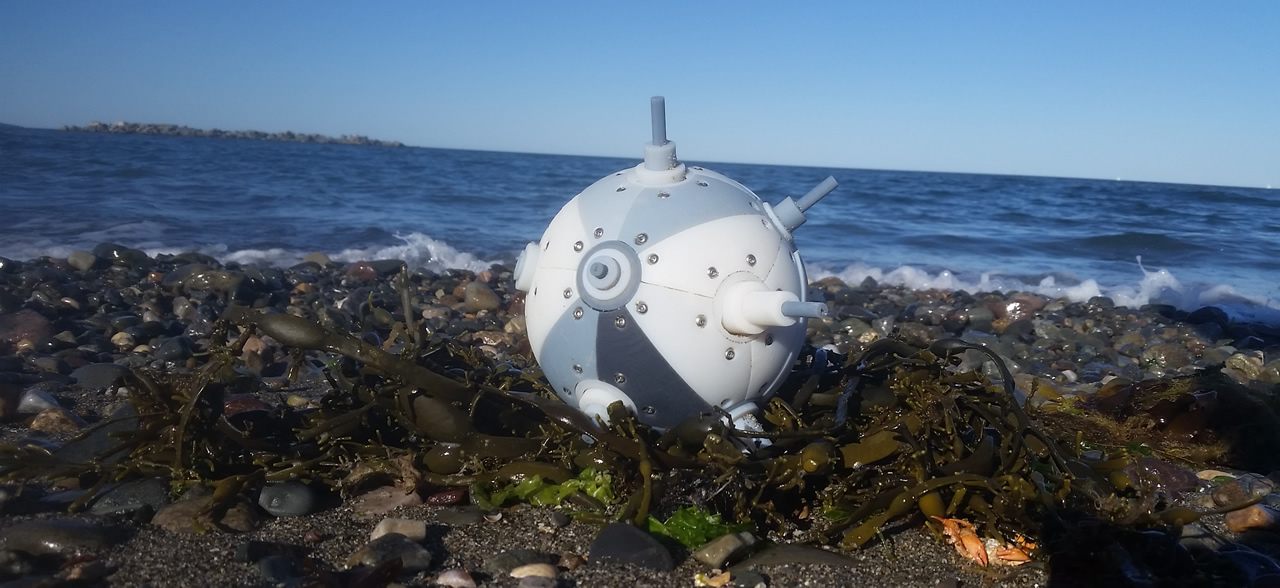}
	\includegraphics[width=0.38\textwidth]{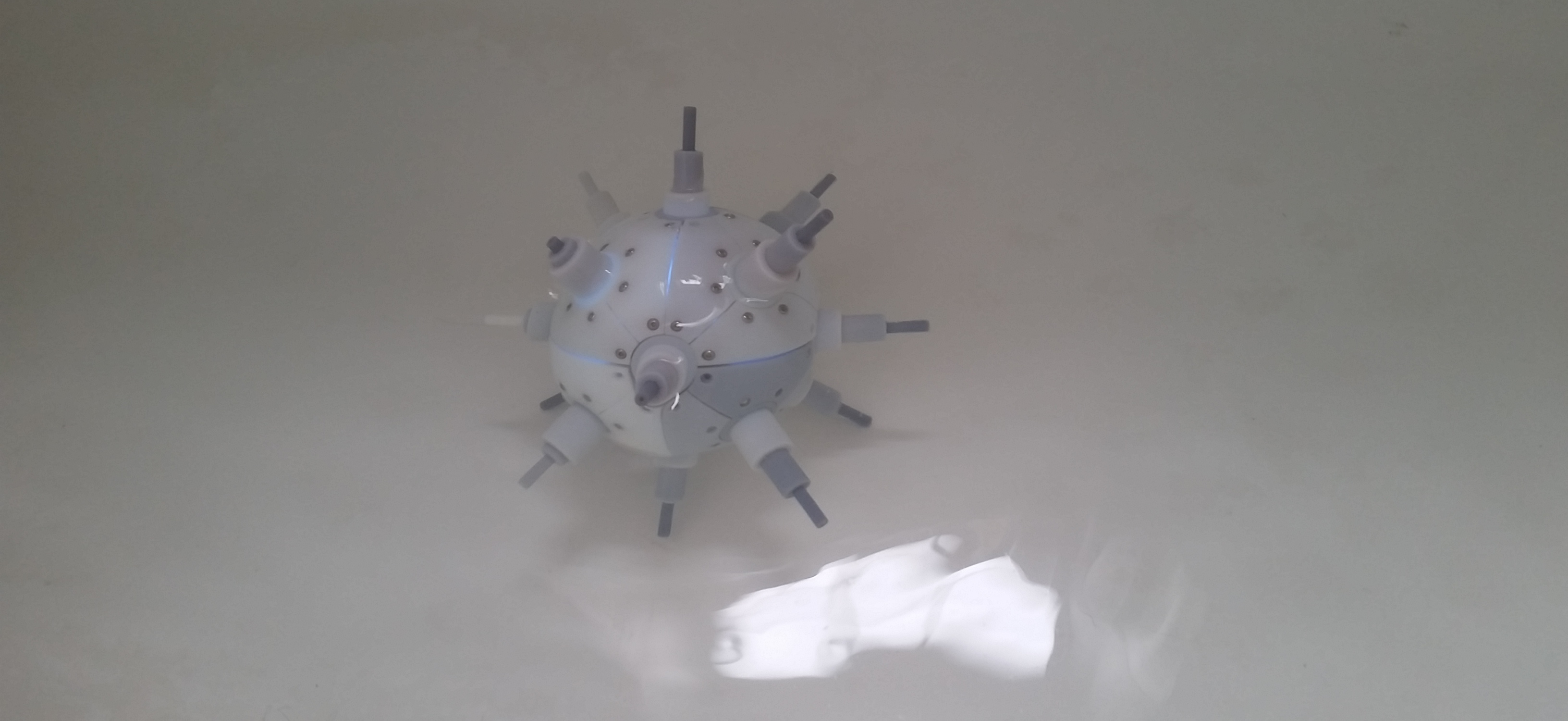}

	\caption{Sea urchin robot moving on snow, sand, rocks and water. }
	\label{figprotosnow}
\end{figure*}

Common spherical mobile robots are sealed and able to float and swim at low speed. If starting from floating static on the water and then rolling constantly to the same direction, then the robot starts to move slowly until it reaches a constant velocity. In contrast, the sea urchin robot is not able to float. It sinks similar to the real sea urchin animals and under the water it can move with its telescopic spines by compressing and extending them in the same fashion as on the ground, but in lower pace, see Fig. \ref{figprotosnow}.


\section{Conclusion and future work}
\label{sec:conclusion}

%

In this paper, a novel telescopic actuation system is presented and integrated in a hybrid bio-inspired sea urchin robot. The telescopic actuators act as the "dynamic" spines of the robot that can be folded inside the spherical body of the robot or can be extended for propulsion purposes.

The actuation system of the telescopic linear actuators consists of linked rack segments that are joined with a hinge and a mechanism for locking/unlocking the segments. In this configuration, when the racks move forward, they are locked and form a rigid rack able to withstand forces in all axes. On the other hand, when the articulated racks are moved backwards they are unlocked and linked with only the hinge, so the articulations can be folded in a minimal space

The robot is bio-inspired from the Echinoid (sea urchin) animal and mimic a sea urchin with the ability to extend and compress its spines. When its spines are contracted, the robot is a sphere and can roll. When its spines are extended in a controlled pattern, it can propel and direct its rolling direction. 

The paper presents a novel actuation system and one of the possible implementations in robotics by integrating it in the bionic sea urchin robot. 


\addtolength{\textheight}{-3cm}   



\bibliographystyle{IEEEtran}
\bibliography{simple}

\end{document}